\documentclass[9pt,technote,compsoc]{IEEEtran}
\ifCLASSOPTIONcompsoc
  \usepackage[nocompress]{cite}
\else
  \usepackage{cite}
\fi
\ifCLASSINFOpdf
  \usepackage[pdftex]{graphicx}
\else
\fi
\ifCLASSOPTIONcompsoc
  \usepackage[caption=false,font=footnotesize,labelfont=sf,textfont=sf]{subfig}
\else
  \usepackage[caption=false,font=footnotesize]{subfig}
\fi
\usepackage{mathtools}
\usepackage{empheq}
\usepackage{amssymb}
\usepackage{amsthm}
\usepackage{pifont}
\usepackage{bm}
\usepackage{booktabs}
\usepackage{adjustbox}
\usepackage{threeparttable}
\usepackage{siunitx}
\usepackage{multirow}
\usepackage[inline]{enumitem}
\usepackage{diagbox}
\usepackage[ruled,vlined]{algorithm2e}
\usepackage[colorlinks]{hyperref}
\usepackage{url}
\usepackage[capitalize]{cleveref}

\newtheorem{theorem}{Theorem}

\DeclareMathOperator*{\argmax}{arg\,max}

\DeclareMathOperator{\Tr}{Tr}
\DeclareMathOperator{\bigO}{\mathcal{O}}
\DeclareMathOperator{\diag}{diag}
\newcommand{\ind}{\mathrm{Ind}}

\newcommand{\colorpm}[1]{_{\pm#1}}

\makeatletter
\DeclareRobustCommand\onedot{\futurelet\@let@token\@onedot}
\def\@onedot{\ifx\@let@token.\else.\null\fi\xspace}

\def\eg{\emph{e.g}\onedot} 
\def\ie{\emph{i.e}\onedot}

\def\wrt{w.r.t\onedot} 
\def\etal{\emph{et al}\onedot}

\makeatother

\begin{document}
%
\title{A Novel Normalized-Cut Solver with Nearest Neighbor Hierarchical Initialization}
%
%
%
%

\author{Feiping~Nie,~\IEEEmembership{Senior~Member,~IEEE,}
	Jitao~Lu,
	Danyang~Wu,
	Rong~Wang,
	and~Xuelong~Li,~\IEEEmembership{Fellow,~IEEE}
	\IEEEcompsocitemizethanks{%
		\IEEEcompsocthanksitem Feiping Nie and Jitao Lu are with the School of
		Computer Science, School of Artificial Intelligence, OPtics and ElectroNics
		(iOPEN), Northwestern Polytechnical University, Xi’an 710072,
		P.R. China, and also with the Key Laboratory of Intelligent Interaction
		and Applications (Northwestern Polytechnical University), Ministry of
		Industry and Information Technology, Xi’an 710072, P.R. China.
		E-mail: \{feipingnie, dianlujitao\}@gmail.com;
		\IEEEcompsocthanksitem Danyang Wu is with the State Key Laboratory for
		Manufacturing Systems Engineering and the School of Electronic and Information
		Engineering, Xi'an Jiaotong University, Xi'an 710049, China. E-mail:
		danyangwu.cs@gmail.com;
		\IEEEcompsocthanksitem Rong Wang and Xuelong Li are with the School of
		Artificial Intelligence, OPtics and ElectroNics (iOPEN), Northwestern
		Polytechnical University, Xi’an 710072, P.R. China, and also with the
		Key Laboratory of Intelligent Interaction and Applications (Northwestern
		Polytechnical University), Ministry of Industry and Information
		Technology, Xi’an 710072, P.R. China. E-mail:
		wangrong07@tsinghua.org.cn; li@nwpu.edu.cn;}%
	\thanks{This work was supported in part by the National Natural Science
		Foundation of China under Grant 62176212.}%
	\thanks{(Corresponding author: Feiping Nie.)}}

%
%

\markboth{Journal of \LaTeX\ Class Files,~Vol.~14, No.~8, August~2015}%
{Shell \MakeLowercase{\textit{et al.}}: Bare Demo of IEEEtran.cls for Computer Society Journals}
%



\IEEEtitleabstractindextext{%
	\begin{abstract}
		Normalized-Cut (N-Cut) is a famous model of spectral clustering. The
		traditional N-Cut solvers are two-stage: 1) calculating the continuous
		spectral embedding of normalized Laplacian matrix; 2) discretization via
		$K$-means or spectral rotation. However, this paradigm brings two vital
		problems: 1) two-stage methods solve a relaxed version of the original problem,
        so they cannot obtain good solutions for the original N-Cut problem; 2) solving
		the relaxed problem requires eigenvalue decomposition, which has $\bigO(n^3)$ time
		complexity ($n$ is the number of nodes). To address the problems, we
		propose a novel N-Cut solver designed based on the famous coordinate descent method.
        Since the vanilla coordinate descent method also has $\bigO(n^3)$ time complexity, we
		design various accelerating strategies to reduce the time complexity to
		$\bigO(|E|)$ ($|E|$ is the number of edges).
        To avoid reliance on random initialization which brings
		uncertainties to clustering, we propose an efficient initialization method
		that gives deterministic outputs. Extensive
		experiments on several benchmark datasets demonstrate that
		the proposed solver can obtain larger objective values of N-Cut, meanwhile achieving
		better clustering performance compared to traditional solvers.
	\end{abstract}

	}

\maketitle

\IEEEdisplaynontitleabstractindextext

%
\IEEEpeerreviewmaketitle

\section{Introduction}\label{sec:introduction}

%
%
%
%

\textsc{Spectral} clustering (SC)~\cite{von2007tutorial} partitions the nodes of a
graph into disjoint clusters so that samples from the same cluster are more
similar to samples from different clusters. The impacts of SC are
far-reaching in clustering, and later graph-based clustering methods are deeply
inspired by SC.
The ancestor works of SC are Ratio-Cut (R-Cut)~\cite{hagen1992new} and
Normalized-Cut (N-Cut)~\cite{shi2000normalized,ng2002spectral}. R-Cut firstly
formulated
clustering as a graph cut problem. It minimizes the ratio
between inter-cluster similarities and cluster sizes to avoid the trivial
solution of min-cut. Unlike R-Cut, N-Cut regularizes the clusters by
their degrees to encourage more balanced clustering assignments and refines
the objective as the ratio between inter-cluster similarities and cluster
degrees. However, the original objectives of R-Cut and N-Cut are NP-hard to
solve.
Therefore, previous works follow a two-stage paradigm,
which solves the relaxed versions of R-Cut and N-Cut and uncovers clustering
results via $K$-means or spectral rotation~\cite{stella2003multiclass}.

\noindent\textit{Related works:}
In summary, the major issues of the two-stage paradigm are:
\begin{enumerate*}
	\item information loss due to relaxation of the original objective;
	\item solving the relaxed objective as an eigenvalue problem is inefficient;
	\item the discrete clustering labels could deviate far from the relaxed
	      embeddings.
\end{enumerate*}
Spectral rotation methods aim at addressing 3, while most existing works aim to
improve the efficiency of 2, \ie, speeding up the computation of eigenvectors.
\cite{DBLP:conf/icml/LinC10,DBLP:conf/icml/BoutsidisKG15} uses power iteration
to approximate the eigenvectors and thus avoid EVD.
\cite{DBLP:conf/ecml/ChenGLCM06} sequentially subsamples the Laplacian matrix
within the iterations of the EVD algorithm to reduce its cost.
\cite{DBLP:conf/kdd/YanHJ09} applies SC to representative points and assigns their
labels to associated samples. \cite{DBLP:journals/pami/FowlkesBCM04} subsamples
the similarity matrix to accelerate EVD. \cite{DBLP:journals/tcyb/CaiC15}
performs singular value decomposition on a landmark-based graph as a substitution
for EVD. \cite{DBLP:conf/icml/TremblayPGV16} generates node embeddings that
approximate the pairwise distances of eigenvectors by filtering random graph
signals. \cite{DBLP:conf/kdd/WuCYXXA18} uses Random Binning features to
approximate the similarity matrix.
Also, a number of methods are designed based on the objectives of SC.
\cite{nie2020self,nie2014clustering,nie2016constrained} combines spectral
embedding learning with graph construction or graph approximation
simultaneously. \cite{DBLP:conf/kdd/SadikajVBP21} proposed a representation
learning method for multi-relational graphs with categorical node attributes.
While efforts have been made to address the second and third issues, the first
has hardly received any attention.

Unlike all the aforementioned methods, our work simultaneously addresses all 3
issues of the two-stage paradigm. We propose a fast heuristic solver to
optimize the original objective of N-Cut directly \emph{without any relaxations
	and approximations}. Hence, this work is orthogonal to SC variants that perform
more approximations and we focus on comparison with the conventional solvers.
Our solver, called Fast-CD, is designed based on the
famous Coordinate Descent method. Via skillful equivalent transformations
and accelerating techniques, the computational cost of Fast-CD is reduced to $\bigO(|E|)$ per
iteration, which is more efficient than relaxed solvers based on EVD.
To assist Fast-CD to converge to better local minimums
without replicating clustering multiple times with distinct random initialization
as the traditional solvers do,
we propose an efficient Nearest Neighbor Hierarchical Initialization (N\textsuperscript{2}HI) method
that leverages the first neighbor relations of the input graph. In experiments,
we mainly compare the proposed Fast-CD solver with the traditional relaxed
solver and spectral rotation. Besides, we also evaluate the clustering
performance compared to some traditional and SOTA graph-based clustering models.
From the results of several benchmarks, we observe that
\begin{enumerate*}
	\item the proposed Fast-CD solver can reach the solution with a lower objective value of
	      the N-Cut model and can achieve better clustering
	      performance;
	\item in some benchmarks, the N-Cut model can achieve SOTA performance via the
	      proposed Fast-CD solver.
\end{enumerate*}
It is worthwhile summarizing the main contributions of this paper as follows:
\begin{itemize}
	\item Different from previous works that solve the
	      N-Cut model via relaxation, we propose a fast heuristic solver based on
	      the coordinate descent method to optimize the objective problem
	      directly with $\bigO(|E|)$ time complexity.
	\item The proposed Fast-CD solver is more efficient than relaxed solvers
	      (based on EVD) in theory and practice.
	\item The proposed Fast-CD solver can achieve the solution with
	      a higher objective value of N-Cut and better clustering performance relative to
	      relaxed solvers.
	\item We propose an efficient initializer N\textsuperscript{2}HI to assist
	      the Fast-CD solver. It outputs deterministic initial cluster assignments
	      and thus effectively avoids randomness in the clustering procedure.
\end{itemize}


\noindent\textit{Notations:}
For any matrix $\mathbf{X}$, $\bm{x}_i$ denotes the $i$-th
column, $\bm{x}^i$ denotes the $i$-th row,  $\lVert \mathbf{X} \rVert_F$
denotes the Frobenius norm, $\Tr(\mathbf{X})$ denotes its trace,
and $\diag(\mathbf{X})$ denotes its diagonal elements. $n$
denotes the number of samples. $c$ denotes the number of clusters. $\mathbf{A}
	\in \mathbb{R}^{n \times n}$ denotes the input similarity matrix, $\mathbf{D}
	\in \mathbb{R}^{n \times n}$ denotes its degree matrix, and $\mathbf{L} =
	\mathbf{D} - \mathbf{A}$ denotes its Laplacian matrix. $\ind$ denotes the set
of cluster indicator matrices with $n \times c$ entries, so $\mathbf{Y} \in \ind$
implies $y_{ij}=1$ if the $i$-th sample belongs to the $j$-th cluster, and $y_{ij}=0$
if not.

\section{Normalized-Cut Revisited}%
\label{sec:ncut}
N-Cut defines the objective value as the sum of ratios between
inter-cluster similarities and cluster degrees:
\begin{align}%
	\label{N-cut-min}
	\min_{\mathbf{Y} \in \ind} \sum_{i=1}^c \frac{\bm{y}_i^T \mathbf{L}
		\bm{y}_i}{\bm{y}_i^T \mathbf{D} \bm{y}_i},
\end{align}
which can be rewritten in matrix form as
\begin{align}%
	\label{N-cut-min-matrix}
	\min_{\mathbf{Y} \in \ind} \Tr\left((\mathbf{Y}^T \mathbf{D}
	\mathbf{Y})^{-\frac{1}{2}} \mathbf{Y}^T \mathbf{L} \mathbf{Y} (\mathbf{Y}^T
	\mathbf{D} \mathbf{Y})^{-\frac{1}{2}} \right).
\end{align}
However, the above equivalent problems are hard to optimize directly due to the
discrete constraint of $\mathbf{Y}$. Hence, SC performs a two-step relaxed
optimization for problem~\eqref{N-cut-min-matrix}. Denoting $\mathbf{F} =
	\mathbf{Y}(\mathbf{Y}^T \mathbf{D} \mathbf{Y})^{-\frac{1}{2}}$, we have
$\mathbf{F}^T \mathbf{D} \mathbf{F} = \mathbf{I}$. SC discards the
discreteness condition of $\mathbf{F}$ and yields
\begin{align}%
	\label{N-cut-min-relax}
	\min_{\mathbf{F}^T \mathbf{D} \mathbf{F} = \mathbf{I}}
	\Tr(\mathbf{F}^T \mathbf{L} \mathbf{F})
	\xRightarrow{\mathbf{H} = \mathbf{D}^{\frac{1}{2}} \mathbf{F}}
	\min_{\mathbf{H}^T \mathbf{H} = \mathbf{I}}
	\Tr(\mathbf{H}^T \mathbf{D}^{-\frac{1}{2}} \mathbf{L}
	\mathbf{D}^{-\frac{1}{2}} \mathbf{H}).
\end{align}
Ng~\etal~\cite{ng2002spectral} applies EVD to the \emph{symmetrically normalized
	Laplacian}
$\mathbf{D}^{-\frac{1}{2}} \mathbf{L}
	\mathbf{D}^{-\frac{1}{2}}$ to solve $\mathbf{H}$, \ie,
$\mathbf{D}^{-\frac{1}{2}} \mathbf{L}
	\mathbf{D}^{-\frac{1}{2}} \bm{h}_i = \lambda_i \bm{h}_i$.
Alternatively, re-substituting $\bm{h}_i = \mathbf{D}^{\frac{1}{2}}
	\bm{f}_i$ to the eigenequation yields $\mathbf{D}^{-1} \mathbf{L}
	\bm{f}_i = \lambda_i \bm{f}_i$, so $\mathbf{F}$ can be obtained by
applying EVD to the \emph{random walk normalized Laplacian}
$\mathbf{D}^{-1} \mathbf{L}$, or by solving the generalized
eigenvalue system $\mathbf{L} \bm{f}_i = \lambda_i \mathbf{D}
	\bm{f}_i$ as proposed in \cite{shi2000normalized}. Finally, the clustering
label matrix $\mathbf{Y}$ can be generated from the learned embedding
$\mathbf{F}$ in different ways. The most popular one is applying $K$-means
clustering to $\mathbf{F}$ or $\mathbf{H}$ to obtain discrete labels. Apart
from $K$-means, spectral rotation methods are also popular for label discretization.
In \cite{stella2003multiclass}, they first calculate an approximation of
$\mathbf{Y}$ via $\mathbf{Y}_0 = \diag\left(
	\diag^{-\frac{1}{2}}(\mathbf{F}\mathbf{F}^T)\right) \mathbf{F}$,
then uncover $\mathbf{Y}$ by solving the following problem:
\begin{align}%
	\label{eq:msc}
	\min_{\mathbf{Y} \in \ind, \mathbf{R}^T \mathbf{R} = \mathbf{I}} \lVert
	\mathbf{Y} - \mathbf{Y}_0 \mathbf{R} \rVert_F^2,
\end{align}
where $\mathbf{R} \in \mathbb{R}^{c \times c}$ is the rotation matrix. In
\cite{huang2013spectral}, the authors compare $K$-means clustering with
spectral rotations in theory and practice. Recently, \cite{chen2017scalable}
argues that solving $\mathbf{Y}$ based on the approximation $\mathbf{Y}_0$
affects the quality of the solution. To alleviate this problem, they propose to
improve problem~\eqref{eq:msc} as
\begin{align}
	\min_{\mathbf{Y} \in \ind, \mathbf{R}^T \mathbf{R} = \mathbf{I}} \lVert
	\mathbf{D}^{\frac{1}{2}} \mathbf{Y} (\mathbf{Y}^T \mathbf{D}
	\mathbf{Y})^{-\frac{1}{2}} - \mathbf{H} \mathbf{R} \rVert_F^2.
\end{align}
In a word, previous works optimize problem~\eqref{N-cut-min} in a two-stage
paradigm:
\begin{enumerate*}
	\item calculating an $n \times c$ continuous embedding $\mathbf{F}$ or $\mathbf{H}$;
	\item learning $\mathbf{Y} \in \ind$ from them with different strategies.
\end{enumerate*}
It is known that the two-stage paradigm generally decreases the quality of
the solution compared to the direct paradigm. In this paper, we focus on optimizing
the objective of N-Cut directly.

\section{Fast Coordinate Descent N-Cut Solver}%
\label{sec:solver}
In this section, we are back to the vector-form~\eqref{N-cut-min} of N-Cut.
Since $\mathbf{L} = \mathbf{D} - \mathbf{A}$, problem~\eqref{N-cut-min} is
equivalent to
\begin{align}%
	\label{N-cut-max}
	\max_{\mathbf{Y} \in \ind} \sum_{i=1}^c \frac{\bm{y}_i^T \mathbf{A}
		\bm{y}_i}{\bm{y}_i^T \mathbf{D} \bm{y}_i}.
\end{align}
Observing each row of $\mathbf{Y}$ being independent, we consider utilizing the
coordinate descent method to directly optimize problem~\eqref{N-cut-max} by
sequentially updating the rows.
Suppose $\mathbf{Y}$ is the current solution and we are updating the
$m$-th row with the rest $\{1, \dots, m-1, m+1, \dots, n\}$-th rows fixed,
there are $c$ alternative choices $\{ \mathcal{Q}_c (k) \}_{k=1}^c$
where $\mathcal{Q}_c (k) \in \mathbb{R}^{c}$ is an one-hot
row vector whose $k$-th element is $1$, and we choose the
one that maximizes \cref{N-cut-max}. For convenience, $\mathcal{Q}_c (0)$ is
the $\mathbb{R}^{c}$ vector with all elements being $0$, and the matrix whose $m$-th
row is $\mathcal{Q}_c (k)$ and other rows are same as $\mathbf{Y}$ is denoted
by $\mathbf{Y}^{(k)}$. Then, the update of $\mathbf{Y}$'s $m$-th row can be
formally expressed as
\begin{align}%
	\label{FCD-1}
	\bm{\hat{y}}^m = \mathcal{Q}_c \left(\argmax_{k \in \{1,\dots,c\}}
	\sum_{i=1}^c \frac{(\bm{y}_i^{(k)})^T \mathbf{A}
	\bm{y}_i^{(k)}}{(\bm{y}_i^{(k)})^T \mathbf{D} \bm{y}_i^{(k)}} \right),
\end{align}
where $\bm{y}_i^{(k)}$ is the $i$-th column of $\mathbf{Y}^{(k)}$, and
$\bm{\hat{y}}^m$ stands for updated $\bm{y}^m$. However,
directly calculating \cref{FCD-1} is expensive because the time complexities of
calculating $\{(\bm{y}_i^{(k)})^T \mathbf{A} \bm{y}_i^{(k)}\}_{k=1}^c$ and
$\{(\bm{y}_i^{(k)})^T \mathbf{D} \bm{y}_i^{(k)}\}_{k=1}^c$
are $\bigO(n^2)$ and $\bigO(n)$. Thus, the time complexity of updating all
$n$ rows adds up to $\bigO(n^3)$, which is the same as EVD.

\subsection{Our Fast Implementation}

To reduce the overhead, we design several strategies to accelerate the
calculation of \cref{FCD-1}. At first, we observe that the calculations for
$(\bm{y}_i^{(k)})^T \mathbf{A} \bm{y}_i^{(k)}$ and $(\bm{y}_i^{(k)})^T
	\mathbf{D} \bm{y}_i^{(k)}$ with different $k$ are redundant since
$\{\mathbf{Y}^{(k)}\}_{k=1}^c$ only differ in $m$-th row. Hence, we consider
transforming \cref{N-cut-max} as follows:
\begin{equation}%
	\label{FCD-2}
	\begin{split}
		& \max_{k \in \{1,\dots,c\}} \sum_{i=1}^c \frac{(\bm{y}_i^{(k)})^T \mathbf{A} \bm{y}_i^{(k)}}{(\bm{y}_i^{(k)})^T \mathbf{D} \bm{y}_i^{(k)}}                                                                                                                  \\
		\xLeftrightarrow{\text{\ding{192}}} & \max_{k \in \{1,\dots,c\}} \sum_{i=1}^c \left[
		\frac{(\bm{y}_i^{(k)})^T \mathbf{A} \bm{y}_i^{(k)}}{(\bm{y}_i^{(k)})^T
		\mathbf{D} \bm{y}_i^{(k)}} - \frac{(\bm{y}_i^{(0)})^T \mathbf{A}
		\bm{y}_i^{(0)}}{(\bm{y}_i^{(0)})^T \mathbf{D} \bm{y}_i^{(0)}} \right] \\
		\xLeftrightarrow{\text{\ding{193}}} & \max_{k \in \{1,\dots,c\}} \mathcal{L}(k) = \frac{(\bm{y}_k^{(k)})^T \mathbf{A} \bm{y}_k^{(k)}}{(\bm{y}_k^{(k)})^T \mathbf{D} \bm{y}_k^{(k)}} -  \frac{(\bm{y}_k^{(0)})^T \mathbf{A} \bm{y}_k^{(0)}}{(\bm{y}_k^{(0)})^T \mathbf{D} \bm{y}_k^{(0)}},
	\end{split}
\end{equation}
where \ding{192} holds because $\frac{(\bm{y}_i^{(0)})^T \mathbf{A}
	\bm{y}_i^{(0)}}{(\bm{y}_i^{(0)})^T \mathbf{D} \bm{y}_i^{(0)}}$ is a constant,
and \ding{193} holds because $\frac{(\bm{y}_i^{(k)})^T \mathbf{A}
	\bm{y}_i^{(k)}}{(\bm{y}_i^{(k)})^T \mathbf{D} \bm{y}_i^{(k)}} = \frac{(\bm{y}_i^{(0)})^T \mathbf{A}
	\bm{y}_i^{(0)}}{(\bm{y}_i^{(0)})^T \mathbf{D} \bm{y}_i^{(0)}}, \forall i\ne k$.
After the above transformations, we avoided calculating $(\bm{y}_i^{(k)})^T
	\mathbf{A} \bm{y}_i^{(k)}$ and $(\bm{y}_i^{(k)})^T \mathbf{D} \bm{y}_i^{(k)},
	\forall i \ne k$, and problem~\eqref{FCD-1} is simplified as
\begin{align} \label{FCD-final}
	\bm{\hat{y}}^m = \mathcal{Q}_c \left(\argmax_{k \in \{1,\dots,c\}}
	\frac{(\bm{y}_k^{(k)})^T \mathbf{A} \bm{y}_k^{(k)}}{(\bm{y}_k^{(k)})^T
	\mathbf{D} \bm{y}_k^{(k)}} -  \frac{(\bm{y}_k^{(0)})^T \mathbf{A}
	\bm{y}_k^{(0)}}{(\bm{y}_k^{(0)})^T \mathbf{D} \bm{y}_k^{(0)}} \right).
\end{align}

The next issue is how to efficiently calculate
$\{(\bm{y}_k^{(k)})^T \mathbf{A} \bm{y}_k^{(k)}\}_{k=1}^c$ and $\{(\bm{y}_k^{(k)})^T
	\mathbf{D} \bm{y}_k^{(k)}\}_{k=1}^c$.  To this goal, we first calculate $\bm{y}_k^T
	\mathbf{A} \bm{y}_k$, $\bm{y}_k^T \mathbf{D} \bm{y}_k$ and $\bm{y}_k^T
	\bm{a}_m$. Then, $(\bm{y}_k^{(k)})^T \mathbf{A} \bm{y}_k^{(k)}$ and
$(\bm{y}_k^{(k)})^T \mathbf{D} \bm{y}_k^{(k)}$ can be inferred from them as the
following two cases supposing the $p$-th element of $\bm{y}^m$ is $1$:

1) If $k = p$, it is easy to observe that $\bm{y}_k^{(k)} = \bm{y}_k$ and
$\bm{y}_k^{(0)} = \bm{y}_k - \mathcal{Q}_n(m)$, where $\mathcal{Q}_n(m) \in
	\mathbb{R}^{n}$ is an one-hot column vector whose $m$-th element is 1. Then the terms
about $\bm{y}_k^{(k)}$ in \cref{FCD-final} can be simply calculated as
\begin{align} \label{FCD-c1-yk}
	\frac{(\bm{y}_k^{(k)})^T \mathbf{A} \bm{y}_k^{(k)}}{(\bm{y}_k^{(k)})^T \mathbf{D} \bm{y}_k^{(k)}} = \frac{\bm{y}_k^T \mathbf{A} \bm{y}_k}{\bm{y}_k^T \mathbf{D} \bm{y}_k}.
\end{align}
And the term about $\bm{y}_k^{(0)}$ in \cref{FCD-final} can be calculated as
\begin{equation}%
	\label{FCD-c1-y0}
	\begin{aligned}
		  & \frac{(\bm{y}_k^{(0)})^T \mathbf{A} \bm{y}_k^{(0)}}{(\bm{y}_k^{(0)})^T \mathbf{D} \bm{y}_k^{(0)}} = \frac{(\bm{y}_k - \mathcal{Q}_n(m))^T \mathbf{A} (\bm{y}_k - \mathcal{Q}_n(m))}{(\bm{y}_k - \mathcal{Q}_n(m))^T \mathbf{D} (\bm{y}_k - \mathcal{Q}_n(m))} \\
		= & \frac{\bm{y}_k^T \mathbf{A} \bm{y}_k - 2\bm{y}_k^T \bm{a}_m + a_{mm}}{\bm{y}_k^T \mathbf{D} \bm{y}_k - 2\bm{y}_k^T \bm{d}_m + d_{mm}}  = \frac{\bm{y}_k^T \mathbf{A} \bm{y}_k - 2\bm{y}_k^T \bm{a}_m}{\bm{y}_k^T \mathbf{D} \bm{y}_k -  d_{mm}},
	\end{aligned}
\end{equation}
where we eliminate $a_{mm}$ in the last step because the diagonal elements of
$\mathbf{A}$ are zeros.  Hence, the objective value becomes
\begin{align}%
	\label{FCD-c1-final}
	\mathcal{L}(k \mid k=p) 	= \frac{\bm{y}_k^T \mathbf{A} \bm{y}_k}{\bm{y}_k^T \mathbf{D} \bm{y}_k} - \frac{\bm{y}_k^T \mathbf{A} \bm{y}_k - 2\bm{y}_k^T \bm{a}_m}{\bm{y}_k^T \mathbf{D} \bm{y}_k -  d_{mm}}.
\end{align}
The time complexity of calculating \cref{FCD-c1-final} is $\bigO(1)$ because
all the terms on the right-hand side have been calculated before.

2) If $k \neq p$,  we can observe that $\bm{y}_k^{(0)} = \bm{y}_k$ and
$\bm{y}_k^{(k)} = \bm{y}_k + \mathcal{Q}_n(m)$. Then the terms about
$\bm{y}_k^{(0)}$ in \cref{FCD-final} can be simply calculated as
\begin{align} \label{FCD-c2-y0}
	\frac{(\bm{y}_k^{(0)})^T \mathbf{A} \bm{y}_k^{(0)}}{(\bm{y}_k^{(0)})^T \mathbf{D} \bm{y}_k^{(0)}} = \frac{\bm{y}_k^T \mathbf{A} \bm{y}_k}{\bm{y}_k^T \mathbf{D} \bm{y}_k}.
\end{align}
And the term about $\bm{y}_k^{(k)}$ in \cref{FCD-final} can be calculated as
\begin{equation} \label{FCD-c2-yk}
	\begin{aligned}
		  & \frac{(\bm{y}_k^{(k)})^T \mathbf{A} \bm{y}_k^{(k)}}{(\bm{y}_k^{(k)})^T \mathbf{D} \bm{y}_k^{(k)}} = \frac{(\bm{y}_k + \mathcal{Q}_n(m))^T \mathbf{A} (\bm{y}_k + \mathcal{Q}_n(m))}{(\bm{y}_k + \mathcal{Q}_n(m))^T \mathbf{D} (\bm{y}_k + \mathcal{Q}_n(m))} \\
		= & \frac{\bm{y}_k^T \mathbf{A} \bm{y}_k + 2\bm{y}_k^T \bm{a}_m + a_{mm}}{\bm{y}_k^T \mathbf{D} \bm{y}_k + 2\bm{y}_k^T \bm{d}_m + d_{mm}}  = \frac{\bm{y}_k^T \mathbf{A} \bm{y}_k + 2\bm{y}_k^T \bm{a}_m}{\bm{y}_k^T \mathbf{D} \bm{y}_k +  d_{mm}}.
	\end{aligned}
\end{equation}
Then the objective value of \cref{FCD-final} \wrt $k \ne p$ becomes
\begin{align}%
	\label{FCD-c2-final}
	\mathcal{L}(k \mid k \ne p) =  \frac{\bm{y}_k^T \mathbf{A} \bm{y}_k + 2\bm{y}_k^T \bm{a}_m}{\bm{y}_k^T \mathbf{D} \bm{y}_k +  d_{mm}} - \frac{\bm{y}_k^T \mathbf{A} \bm{y}_k}{\bm{y}_k^T \mathbf{D} \bm{y}_k} .
\end{align}
The time complexity of calculating \cref{FCD-c2-final} is $\bigO(1)$ because
all the terms on the right-hand side have been calculated before. Traversing
$k$ from $1$ to $c$, the total time complexity is $\bigO(c)$.
After obtaining all $\{\mathcal{L}(k)\}_{k=1}^c$, we update $\bm{\hat{y}}^m$ to
the one that maximizes it.

Since \cref{FCD-c1-final,FCD-c2-final} are efficiently calculated based on
previously known $\bm{y}_k^T \mathbf{A} \bm{y}_k, \bm{y}_k^T \mathbf{D} \bm{y}_k$
and $\bm{y}_k^T \bm{a}_m$, we need to refresh them before updating the
next row of $\mathbf{Y}$. Suppose $r = \argmax_k \mathcal{L}(k)$,
the solution remains unchanged if $r=p$ (recall that $p$ stands for the index
of $1$ prior to updating $\bm{y}^m$), so there's no need to update them.
Otherwise, we consider the following two additional cases:

\noindent 1) For variables involving $\bm{y}_r$, we have $\bm{\hat{y}}_r = \bm{y}_r^{(r)}$ so
they can be calculated as
\begin{equation}%
	\label{eq:update_new}
	\begin{aligned}
		\bm{\hat{y}}_r^T \mathbf{A} \bm{\hat{y}}_r & = (\bm{y}_r^{(r)})^T \mathbf{A}
		\bm{y}_r^{(r)}, \quad \bm{\hat{y}}_r^T \mathbf{D} \bm{\hat{y}}_r =
		(\bm{y}_r^{(r)})^T \mathbf{D} \bm{y}_r^{(r)},                                              \\
		\bm{\hat{y}}_r^T \mathbf{A}                  & = \bm{y}_r^T \mathbf{A} +
		\bm{a}^{m}.
	\end{aligned}
\end{equation}
2) For variables involving $\bm{y}_p$, we have $\bm{\hat{y}}_p =
	\bm{y}_p^{(0)}$ then they can be calculated as
\begin{equation}%
	\label{eq:update_old}
	\begin{aligned}
		\bm{\hat{y}}_p^T \mathbf{A} \bm{\hat{y}}_p & = (\bm{y}_p^{(0)})^T \mathbf{A}
		\bm{y}_p^{(0)}, \quad \bm{\hat{y}}_p^T \mathbf{D} \bm{\hat{y}}_p =
		(\bm{y}_p^{(0)})^T \mathbf{D} \bm{y}_p^{(0)},                                              \\
		\bm{\hat{y}}_p^T \mathbf{A}                  & = \bm{y}_p^T \mathbf{A} -
		\bm{a}^{m}.
	\end{aligned}
\end{equation}
Obviously, calculating $\{\bm{\hat{y}}_k^T \mathbf{A} \bm{\hat{y}}_k,
	\bm{\hat{y}}_k^T \mathbf{D} \bm{\hat{y}}_k \mid k \in \{r, p\}\}$ does not
introduce extra complexity
since all the variables have been calculated in previous steps. Calculating
$\{\bm{\hat{y}}_k^T \mathbf{A} \mid k \in \{r, p\}\}$
is only necessary after actually updating $\mathbf{Y}$ (\ie, $r \ne p$). The whole procedure is summarized in
\cref{alg:fast_cd}. We break the outer iteration when the increasing rate of
objective value is lower than a threshold, which is $10^{-9}$ throughout the
experiments.

\begin{algorithm}[t]
	\KwIn{$\mathbf{A} \in \mathbb{R}^{n \times n}, \mathbf{Y} \in \mathbb{R}^{n \times c}$}
	Calculate and store $\bm{y}_k^T \mathbf{A}$, $\mathbf{D}$, $\bm{y}_k^T
	\mathbf{A} \bm{y}_k$, $\bm{y}_k^T \mathbf{D}
	\bm{y}_k$, and $n_k = \bm{y}_k^T\bm{1}$, $\forall k \in \{1,\dots,c\}$\;
	\For(\tcp*[f]{outer iteration}){$t \gets 1$ \KwTo $T$}{
	\For(\tcp*[f]{inner iteration}){$m \gets 1$ \KwTo $n$}{
	$p \gets$ the index of element $1$ in $\bm{y}^m$\;
	\If(\tcp*[f]{Avoid empty cluster}){$n_p = 1$}{
		continue\;
	}
	Calculate $\mathcal{L}(k), \forall k$ via
	\cref{FCD-c1-final} or \cref{FCD-c2-final}\;
	$r \gets \argmax_{k \in \{1,\dots,c\}} \mathcal{L}(k)$\;
	\If(\tcp*[f]{Better solution found}){$r \ne p$}{
	Update $\bm{y}_r^T \mathbf{A} \bm{y}_r, \bm{y}_r^T \mathbf{D}
		\bm{y}_r$, $\bm{y}_r^T \mathbf{A}$ via \cref{eq:update_new}\;
	Update $\bm{y}_p^T \mathbf{A} \bm{y}_p, \bm{y}_p^T \mathbf{D}
		\bm{y}_p$, $\bm{y}_p^T \mathbf{A}$ via \cref{eq:update_old}\;
	$y_{mr} \gets 1, y_{mp} \gets 0$,
		$n_r \gets n_r + 1, n_p \gets n_p - 1$\;
	}
	}
	}
	\KwOut{$\mathbf{Y}$}
	\caption{Fast-CD solver for N-Cut problem~\eqref{FCD-2}}%
	\label{alg:fast_cd}
\end{algorithm}

\subsection{Complexity Analyses}%
\label{ssec:comp_ana}

In real-world scenarios, the input graph is usually sparse.
For convenience, we assume $\mathbf{A}$ has $s$ non-zero entries in each row
so the number of edges $|E| = ns$.

\noindent\textit{Time complexity:}
Since $\mathbf{Y}$ contains exact $n$ non-zero elements (one in each row),
it's highly sparse and related variables can be efficiently computed.
\begin{enumerate*}
	\item In the initialization stage,
$\{\bm{y}_k^T \mathbf{A} \}_{k = 1}^c$ are the sums of edge weights associated
with each cluster, so calculating them needs $ns$ additions in total. Then,
calculating $\mathbf{D}$ by $d_{ii} = \sum_k \bm{y}_k^T \bm{a}_i$ needs at most
$n\min(s,c)$ additions, because it's the sum of edge
weights from node $i$ to nodes in cluster $k$, and each sample can connect to
at most $\min(s,c)$ clusters.
Calculating $\{\bm{y}_k^T \mathbf{A} \bm{y}_k\}_{k=1}^c$ based on known
$\{\bm{y}_k^T \mathbf{A} \}_{k = 1}^c$ needs $n$ additions.
$\{\bm{y}_k^T \mathbf{D} \bm{y}_k\}_{k=1}^c$ can be obtained by summing
$d_{ii}$ for each cluster, which needs $n$ additions.
Hence, the time complexity of this stage is $\bigO(ns)$.
\item
Within each \emph{inner iteration}, calculating
\cref{FCD-c1-final} or \cref{FCD-c2-final} needs $3c$ additions/subtractions
and $2c$ divisions, whose time complexity is $\bigO(c)$.
\item
If a better solution is found, updating stored
variables as \cref{eq:update_new,eq:update_old}
needs $2s$ additions/subtractions with time complexity $\bigO(s)$.
\end{enumerate*}
Denote the number of \emph{outer iterations}
and the number of \emph{better solution found} as $T$ and $n_t$, the overall time complexity of
Fast-CD can be expressed as $\bigO(ns + Tnc + \sum_{t=1}^T n_t s)$, where $n_t \leqslant n$
and $T \ll n$. Also, $n_t$ decreases in each iteration and
eventually becomes $0$, so $\sum_{t} n_t < Tn$. In the worst case where $\mathbf{A}$ is dense,
calculating $\{\bm{y}_k^T \mathbf{A} \}_{k = 1}^c$ needs $n^2$ additions, $\mathbf{D}$ needs
$nc$ additions, and updating stored $\bm{y}_r^T \mathbf{A}$, $\bm{y}_p^T
\mathbf{A}$ needs $2n$ additions, so the time complexity becomes $\bigO(n^2 +
Tnc + \sum_{t=1}^T n_t n)$ and still more efficient than $\bigO(n^3)$ EVD.

\noindent\textit{Space complexity:}
For the intrinsic variables of the N-Cut problem, storing $\mathbf{A}$ takes
$\bigO(ns)$ if sparse and $\bigO(n^2)$ if dense, both of $\mathbf{Y}$ and
$\mathbf{D}$ take $\bigO(n)$.  Additionally, Fast-CD maintains $3c$ variables
$\bm{y}_k^T \mathbf{A} \bm{y}_k$, $\bm{y}_k^T \mathbf{D} \bm{y}_k$, $n_k$ and
at most $n\min(s,c)$ variables $\bm{y}_k^T \mathbf{A}$ in sparse case or exact
$nc$ variables in dense case.  Thus, its space complexity is $\bigO(ns)$ if
sparse and $\bigO(n^2)$ if dense.

\subsection{Discussion}%

\noindent\textit{Convergence of Fast-CD:}
\Cref{alg:fast_cd} is guaranteed to converge according to the following
theorem.
\begin{theorem}
	Given initial cluster assignments, \cref{alg:fast_cd} monotonically increases
	the corresponding N-Cut objective~\eqref{N-cut-max} along with each iteration
	until convergence.
\end{theorem}
\begin{proof}
	Denote the solution prior to the $m$-th inner iteration as
	$\mathbf{Y}$ where $\bm{y}^{m}=\mathcal{Q}_c(p)$. After the $m$-th inner
	iteration, we obtain $\hat{\mathbf{Y}}$ by solving
	problem~\eqref{FCD-final} where $\hat{\bm{y}}^{m}=\mathcal{Q}_c(r)$, then
	we have $\mathcal{L}(p) \leqslant \mathcal{L}(r)$
	and the equality holds if $r=p$. Thus, according to
	\cref{FCD-2}, the objective of N-Cut will also increase.
\end{proof}

\noindent\textit{Avoid empty clusters:}
It's obvious that problem~\eqref{N-cut-max} is equivalent to $\max_{\mathbf{Y}
\in \ind} \sum_{i=1}^c \frac{\bm{y}_i^T (\lambda\mathbf{D} + \mathbf{A})
\bm{y}_i}{\bm{y}_i^T \mathbf{D} \bm{y}_i}$. Compared to
problem~\eqref{N-cut-max}, each cluster's objective
increases $\lambda$ if non-empty and remains $0$
if empty ($\bm{y}_i = \bm{0}$). When $\lambda$ is large enough, the whole
objective of all clusters being non-empty is guaranteed to be larger than that
of any cluster being empty, so solutions with empty clusters won't be picked.
We have the following theorem for the concrete value of $\lambda$:
\begin{theorem}
	Solving the above problem by Fast-CD won't produce empty
	clusters when $\lambda\mathbf{D} + \mathbf{A}$ is positive semi-definite
	(PSD), where $\lambda$ is a constant to ensure positive semi-definiteness.
\end{theorem}
\begin{proof}
	When PSD, it can be decomposed as $\lambda\mathbf{D} + \mathbf{A} =
	\mathbf{P}^T\mathbf{P}$, and
	Fast-CD objective~\eqref{FCD-2} for the above problem becomes
	\begin{equation}
		\label{eq:CD_lam}
			\max_{k \in \{1,\dots,c\}} \mathcal{J}(k) = \frac{(\bm{y}_k^{(k)})^T \mathbf{P}^T\mathbf{P}
			\bm{y}_k^{(k)}}{(\bm{y}_k^{(k)})^T \mathbf{D} \bm{y}_k^{(k)}} -
			\frac{(\bm{y}_k^{(0)})^T \mathbf{P}^T\mathbf{P}\bm{y}_k^{(0)}}{(\bm{y}_k^{(0)})^T
			\mathbf{D} \bm{y}_k^{(0)}}.
	\end{equation}
If a cluster $\mathcal{C}_p$ contains exactly one sample and it's the $m$-th,
\ie, $\bm{y}_p^{(p)} = \mathcal{Q}_n(m)$ and $\bm{y}_p^{(0)} = \bm{0}$,
define $\frac{0}{0}=0$, the objectives are
\begin{subequations}
	\begin{empheq}[left={%
			\begin{aligned}[b]&\mathrlap{\mathcal{J}(k)=}\\
		 &\empheqlbrace
		\end{aligned}%
	}]{alignat=2}
			&\textstyle\frac{\bm{p}_m^T\bm{p}_m}{d_{mm}}, &&\text{ if } k = p,
			\label{eq:obj_psd_p}
			\\
			&\textstyle\frac{\bm{y}_k^T \mathbf{P}^T\mathbf{P} \bm{y}_k + 2\bm{y}_k^T\mathbf{P}^T
			\bm{p}_m + \bm{p}_m^T\bm{p}_m}{\bm{y}_k^T \mathbf{D} \bm{y}_k +  d_{mm}} -
			\frac{\bm{y}_k^T \mathbf{P}^T\mathbf{P}\bm{y}_k}{\bm{y}_k^T \mathbf{D} \bm{y}_k},
			&&\text{ otherwise},
			\label{eq:obj_psd_k}
	\end{empheq}
\end{subequations}
in which \cref{eq:obj_psd_k} is obtained similar to \cref{FCD-c2-final}.
With some trivial algebra, subtracting \eqref{eq:obj_psd_k} from
\eqref{eq:obj_psd_p} yields
\begin{equation}
\mathcal{J}(p) - \mathcal{J}(k)=
\frac{\lVert \bm{y}_k^T \mathbf{D} \bm{y}_k \bm{p}_m - d_{mm} \mathbf{P}\bm{y}_k \rVert_2^2}
{d_{mm} \bm{y}_k^T \mathbf{D} \bm{y}_k (d_{mm} + \bm{y}_k^T \mathbf{D} \bm{y}_k)}
\geqslant 0, \forall k\ne p,
\end{equation}
so $\mathcal{C}_p$ won't become empty since leaving it as-is is optimal.
\end{proof}
We simply skip singleton clusters in practice. This is
efficient and does not introduce extra computational complexity, because we can
count each cluster and save them as $n_1, n_2, \dots, n_c$ in the
initialization step. After that, querying and updating the capacity of each
cluster are both $\bigO(1)$.

\noindent\textit{Estimate the number of clusters:}
\begin{figure}[t]
	\centering
	\subfloat[Synthetic 5-block dataset with uniform
	noise.]{\includegraphics[width=0.38\linewidth]{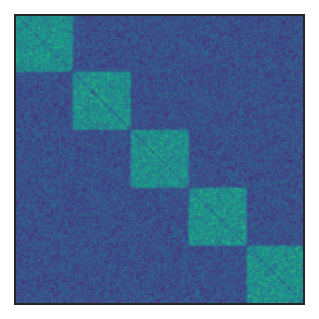}}
	\hspace{2em}
	\subfloat[N-Cut objectives of different numbers of
	clusters.]{\includegraphics[width=0.44\linewidth]{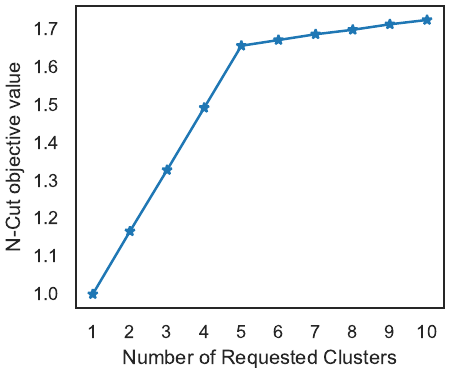}}
	\caption{An illustrative example to estimate the number of clusters $c$ via the
		proposed N-Cut objective gap heuristic. The turnaround point is observed at
		$c=5$ for the synthetic dataset with 5 clusters.}%
	\label{fig:ncut_obj}
\end{figure}
The eigengap heuristic (Section~8.3 of \cite{von2007tutorial}) is particularly
designed for spectral clustering to estimate the number of desired clusters
$c$. Denoting $\lambda_1, \lambda_2, \dots, \lambda_n$ as the eigenvalues of
normalized Laplacian matrix in ascending order, the data is likely to have $c$
clusters if $\lambda_1, \dots, \lambda_c$ are very small but $\lambda_{c+1}$ is
relatively large.
However, the eigengap heuristic is not applicable to our proposed Fast-CD
because we don't do EVD. To address the lack, we propose a heuristic
based on the gap of N-Cut objective \eqref{N-cut-max}. The first step is to
repeatedly apply \cref{alg:fast_cd} with an ascending list of the number of
clusters $[c_1, c_2, \dots, c_\tau]$ around the roughly estimated one, then
calculate N-Cut objectives $[\mathcal{J}_1, \mathcal{J}_2, \dots,
			\mathcal{J}_\tau]$ via \cref{N-cut-max} from corresponding clustering labels.
Finally, we choose the number $c_i$ if $\mathcal{J}_i - \mathcal{J}_{i - 1}$ is
large but $\mathcal{J}_{i + 1} - \mathcal{J}_i$ is relatively small.
\cref{fig:ncut_obj} provides an example of a 5-block dataset, where each block
contains 100 samples and uniform noise is added to bring some difficulty. By
inspecting the turnaround point of the N-Cut objective curve, the proposed
heuristic successfully uncovered the ground-truth number of clusters $c=5$.
In essence, \cref{N-cut-max} measures the sum of \emph{densities} of
resulting clusters.
Owing to the fact that intra-cluster edges are dense while inter-cluster edges
are sparse, \cref{N-cut-max} evaluates to a small value when they contain samples from
more than one ground-truth clusters ($c_i < c$), and saturates when $c_i = c$.
As $c_i$ becomes larger, a ground-truth cluster is likely to be broken into
denser small clusters, so \cref{N-cut-max} will keep increasing but the rate of
the increment will be small. The overall time complexity of the heuristic is
$\bigO((ns + Tnc + \sum_{t=1}^T n_t s)\tau)$, and is more efficient than the $\bigO(n^3)$
eigengap heuristic considering $\tau \ll n$.

\section{Nearest Neighbor Hierarchical Initialization}%
\label{sec:n2hi}

As introduced in \cref{sec:ncut}, $K$-means and spectral rotation are common
discretization strategies for N-Cut. However, a good initialization is crucial
for both of them to obtain satisfactory performance. To be specific, it's
well-known that Lloyd's algorithm for $K$-means is sensitive to the initial
guesses for the clustering centroids and often gets stuck in bad local minima.
There's also no easy way to initialize the rotation matrix $\mathbf{R}$ for
spectral rotation and the initial labels $\mathbf{Y}$ for our Fast-CD solver. A
workaround is to replicate the process multiple times with distinct initial
guesses and pick the one with the best objective value, but the shortcoming is
obvious: it's computationally expensive and brings uncertainties to clustering.

\subsection{Our Methodology}

\begin{figure}[t]
	\centering
	\includegraphics[width=0.7\linewidth]{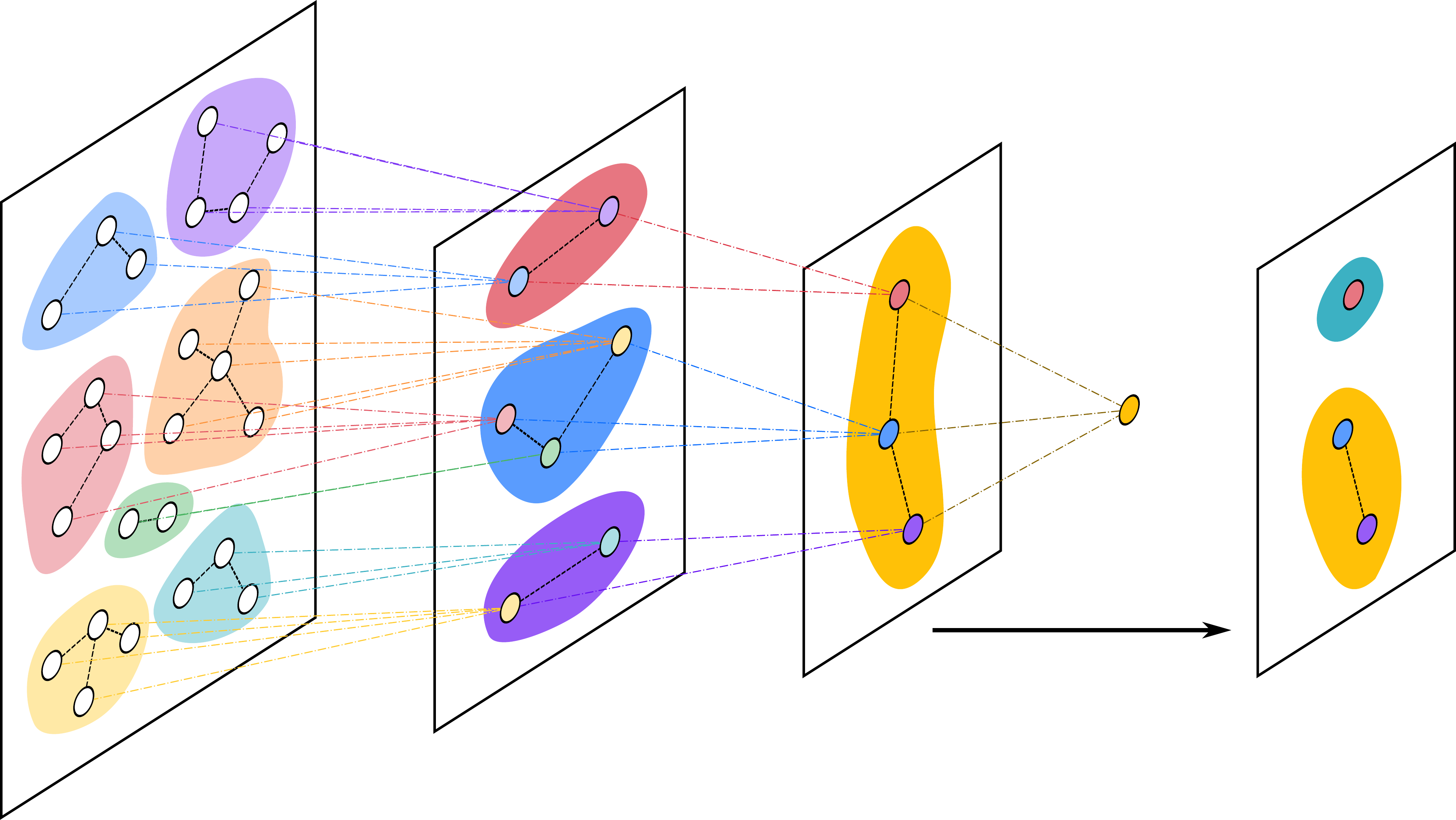}
	\caption{The overview of N\textsuperscript{2}HI.
		Black dashed lines within each level illustrate 1-nn relationships.
		Colored dashed lines across
		neighboring levels illustrate graph coarsening processes.
		A hierarchy of 7--3--1 clusters is first generated based on 1-nearest neighbors.
		The 3-cluster partition is later refined to obtain a 2-cluster partition.}%
	\label{fig:n2hi}
\end{figure}

The nearest neighbor method is one of the most simple yet effective methods for
classification. Each data sample is assigned to the class of its closest
neighbor. The idea is also adopted by the assignment step of $K$-means
clustering, where each data sample is assigned to its nearest cluster centroid.
Recently, Sarfraz~\etal proposed an efficient parameter-free method for data
clustering using first neighbor relations~\cite{sarfraz2019efficient} that
iteratively merges data samples with their 1-nearest neighbor (1-nn) and
generates a cluster hierarchy. Motivated by the simpleness and effectiveness of
nearest neighbor methods, we propose a nearest neighbor hierarchical
initialization (N\textsuperscript{2}HI) method for graph data.
N\textsuperscript{2}HI is efficient and able to produce semantically meaningful
cluster assignments, well coupled with our Fast-CD solver, and always gives
deterministic outputs thus effectively avoiding uncertainties in clustering
tasks. Concretely, N\textsuperscript{2}HI is composed of three steps as
illustrated	in \cref{fig:n2hi}.

\noindent\textit{a) Clustering by first-neighbor relations:}
We denote the input similarity graph at layer $\ell$ as
$\mathbf{A}^{(\ell)}\in\mathbb{R}^{n_\ell \times n_\ell}$, the 1-nn of each data sample
can be obtained by
\begin{align}%
	\label{eq:n2hi_max}
	u_i^{(\ell)}=\argmax_i \bm{a}_i^{(\ell)}, \forall i\in\{1,\dots,n_\ell\}.
\end{align}
Then, we obtain the partition at layer $\ell$ by assigning the $i$- and $j$-th
samples to the same cluster if $i=u^{(\ell)}_j$ or $j=u^{(\ell)}_i$, \ie, one
is the 1-nn of the other.  All data samples and their 1-nn form a bipartite
graph, so the cluster assignments can be obtained by Tarjan's
algorithm~\cite{tarjan1972depth} in linear time to find its connected
components. The input graph at the first layer is $\mathbf{A}^{(1)}=\mathbf{A}$.

\noindent\textit{b) Graph coarsening:}
After computing the cluster assignments, we merge data samples within the same
cluster to obtain the input of the next layer. However, we have no access to
the raw data samples in the context of graph-based clustering, so it's
impossible to directly average them to obtain the representation after merging
to compute new similarities. Instead, we propose to approximate the
inter-cluster similarities of the next layer as
\begin{align}%
	\label{eq:n2hi_merge}
	\mathbf{A}^{(\ell+1)}=\frac{1}{(\mathbf{Y}^{(\ell)})^T\bm{11}^T\mathbf{Y}^{(\ell)}}
	\circ (\mathbf{Y}^{(\ell)})^T \mathbf{A^{(\ell)}} \mathbf{Y}^{(\ell)},
\end{align}
where $\mathbf{Y}^{(\ell)}$ denotes the cluster indicator matrix obtained by
partitioning $\mathbf{A}^{(\ell)}$, $\bm{1}$ is an all-one vector with proper
dimension, $\circ$ denotes the Hadamard product. We provide a 2-cluster example
to facilitate interpreting how \cref{eq:n2hi_merge} works. Given two clusters
$\mathcal{C}_1$, $\mathcal{C}_2$ associated with a similarity matrix
$\mathbf{A}$ and cluster indicator matrix $\mathbf{Y}$, we propose to express
the similarity between a sample $i \in \mathcal{C}_1$ and the whole
$\mathcal{C}_2$ by the averaged similarity between $i$ and all $j \in
	\mathcal{C}_2$:
\begin{align}
	\hat{a}_{i, \mathcal{C}_2} = \frac{1}{|\mathcal{C}_2|} \sum_{j \in
		\mathcal{C}_2} a_{ij},
\end{align}
and vice-versa for $\mathcal{C}_1$ and any $j \in \mathcal{C}_2$. Thus, the
inter-cluster similarity between $\mathcal{C}_1$ and $\mathcal{C}_2$ is the
averaged sum of pairwise similarities of all samples from the two clusters:
\begin{align}
	\hat{a}_{\mathcal{C}_1,\mathcal{C}_2}
	= \frac{1}{|\mathcal{C}_1| \cdot |\mathcal{C}_2|}
	\sum_{i \in \mathcal{C}_1, j \in \mathcal{C}_2} a_{ij}
	= \frac{\bm{y}_{1}^T\mathbf{A} \bm{y}_{2}}
	{\bm{y}_1^T \bm{1} \cdot \bm{y}_2^T \bm{1}},
\end{align}
which is indeed the individual elements of \cref{eq:n2hi_merge}. After
obtaining $\mathbf{A}^{(\ell + 1)}$, it's feed into \cref{eq:n2hi_max} and
the process is repeated until there's only one cluster. The output is a cluster
hierarchy $\mathcal{T}=\{\Gamma_1, \Gamma_2, \dots, \Gamma_L \mid |\Gamma_\ell|
	> |\Gamma_{\ell + 1}|, \forall \ell\}$, in which $\Gamma_\ell =
	\{\mathcal{C}_1, \mathcal{C}_2, \dots, \mathcal{C}_{|\Gamma_\ell|}\}$ are the
sets of clusters.

\noindent\textit{c) Refinement:}
In order to obtain the clustering result with exact $c$ clusters, we provide a
mechanism to refine the hierarchy. If $\exists \Gamma_\ell \in \mathcal{T},
	|\Gamma_\ell|=c$, the result is directly obtained without any subsequent steps.
Otherwise, we find the $\Gamma_\ell$ from $\mathcal{T}$ that satisfying
$|\Gamma_\ell| > c > |\Gamma_{\ell+1}|$, and refine it (essentially we can pick
any $\Gamma_\ell$ as long as $|\Gamma_\ell| > c$). The refinement is composed
of $|\Gamma_\ell| - c$ iterations, in which we subsequently merge one pair
$(u,v)$ with the largest similarity at a time and update the similarities
by $a_{iu} \gets (a_{iu} + a_{iv}) / 2, \forall i\notin\{u,v\}$. Obviously, $c$
cannot be larger than the first partition $|\Gamma_1|$, but we
empirically find that $|\Gamma_1|$ is \emph{always} larger than the ground-truth
$c$ we desire.

\subsection{Complexity Analyses}

At \textit{step a)}, finding 1-nn and obtaining the cluster assignments are both
$\bigO(|E|^{(\ell)})$, where $|E|^{(\ell)}$ denotes the number of edges of the
graph associated with $\mathbf{A^{(\ell)}}$. At \textit{step b)},
\cref{eq:n2hi_merge} can be efficiently calculated in $\bigO(|E|^{(\ell)})$. At
\textit{step c)}, $|\Gamma_\ell|$ clusters are recursively refined to obtain
$|\Gamma_\ell - 1|, |\Gamma_\ell - 2|, \dots, c$ clusters, while each sub-step
needs $|\Gamma_\ell| - 2$ calculations. Considering $|\Gamma_\ell| \ll n$, the
overall time complexity of N\textsuperscript{2}HI is $\bigO(|E|)$.

\subsection{Relationships to Graph Pooling}

Interestingly, we find that our proposed N\textsuperscript{2}HI is similar in
spirit to the graph pooling technique~\cite{mesquita2020rethinking} of a myriad
of graph neural network (GNN)
architectures~\cite{defferrard2016convolutional,ying2018hierarchical,ma2019graph,yuan2019structpool,bianchi2020spectral,gao2021topology},
in which a hierarchy of coarsened graphs is produced
according to node clustering assignments.
We argue that N\textsuperscript{2}HI differs from graph
pooling in the following aspects:
\begin{enumerate*}
	\item graph pooling aims to enhance graph representation learning, while
	      N\textsuperscript{2}HI aims to obtain semantically meaningful cluster
	      assignments;
	\item state-of-the-art graph pooling modules are differentiable and
	      integrated with GNN in an end-to-end manner, \eg,
	      \cite{ying2018hierarchical} uses soft labels learned by GNN to produce the
	      coarsened graph. On the contrary, N\textsuperscript{2}HI leverages 1-nn
	      information from graphs and thus not differentiable. There's no learning,
	      either;
	\item graph embeddings obtained from the last layer of graph pooling module
	      are sent to downstream tasks (\eg, node classification), while we refine an
	      intermediate output of N\textsuperscript{2}HI to initialize subsequent
	      clustering tasks.
\end{enumerate*}

\section{Experiments}%
\label{sec:exp}

In this section, we conduct extensive experiments on different tasks to
demonstrate the superiority of Fast-CD compared with other N-Cut solvers. In
addition, we compare N-Cut with other popular graph-based clustering methods
and discuss the trade-off of choosing the appropriate model for particular
tasks.

\subsection{Evaluation Protocol}

\noindent\textit{Competitors:}
We focus on comparing the proposed Fast-CD with two representative N-Cut
solvers:
\begin{enumerate*}
	\item the most popular Ng-Jordan-Weiss algorithm~\cite{ng2002spectral}
	      (denoted as EVD+KM), which utilizes EVD to compute spectral embeddings and
	      applies $K$-means to obtain labels;
	\item Improved Spectral Rotation (ISR), an N-Cut solver proposed
	      recently~\cite{chen2017scalable} that improves the spectral rotation
	      strategy~\cite{stella2003multiclass}.
\end{enumerate*}

\noindent\textit{Criterion:}
The competitors are essentially different approaches to solving the same
optimization problem~\eqref{N-cut-max}, so justifiably the most important
criterion is the objective value, referred to as \texttt{obj}. After
convergence, the larger the \texttt{obj} is, the better performance the solver has.

\noindent\textit{Settings:}
Apart from the proposed Fast-CD solver, we initialize all methods that
require initial labels with the proposed N\textsuperscript{2}HI method for a
fair comparison. In addition,
the initial $K$-means centroids of EVD+KM and the rotation matrix of ISR are also generated from N\textsuperscript{2}HI labels.
A side effect of this setting is that the clustering results are deterministic across different runs.
All comparison models are implemented in MATLAB R2020b and launched on a
desktop with Intel i7 7700k @ 4.2GHz CPU and 32 GB RAM running Arch Linux
x86\_64.

\subsection{Qualitative Results}

To give an intuitive understanding of the advantages of Fast-CD, we first
compare different N-Cut solvers on a synthetic dataset and the image
segmentation task.

\subsubsection{Synthetic Data}

\begin{figure}[t]
	\centering
	\subfloat[50 noise points. All competitors correctly cluster two circles and converge to the same solution.]{\includegraphics[width=\linewidth]{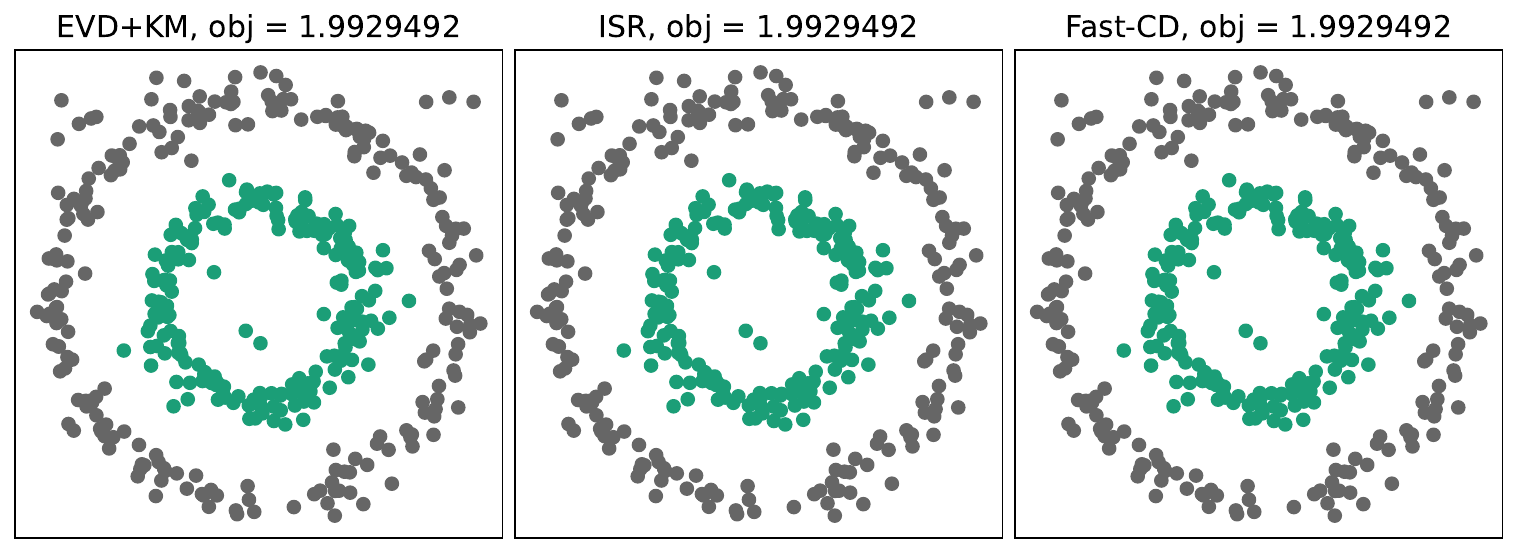}}
	\\
	\subfloat[100 noise points. EVD+KM and ISR fail to cluster the outer
		circle, Fast-CD correctly clusters two circles and achieves the best N-Cut objective.]
	{\includegraphics[width=\linewidth]{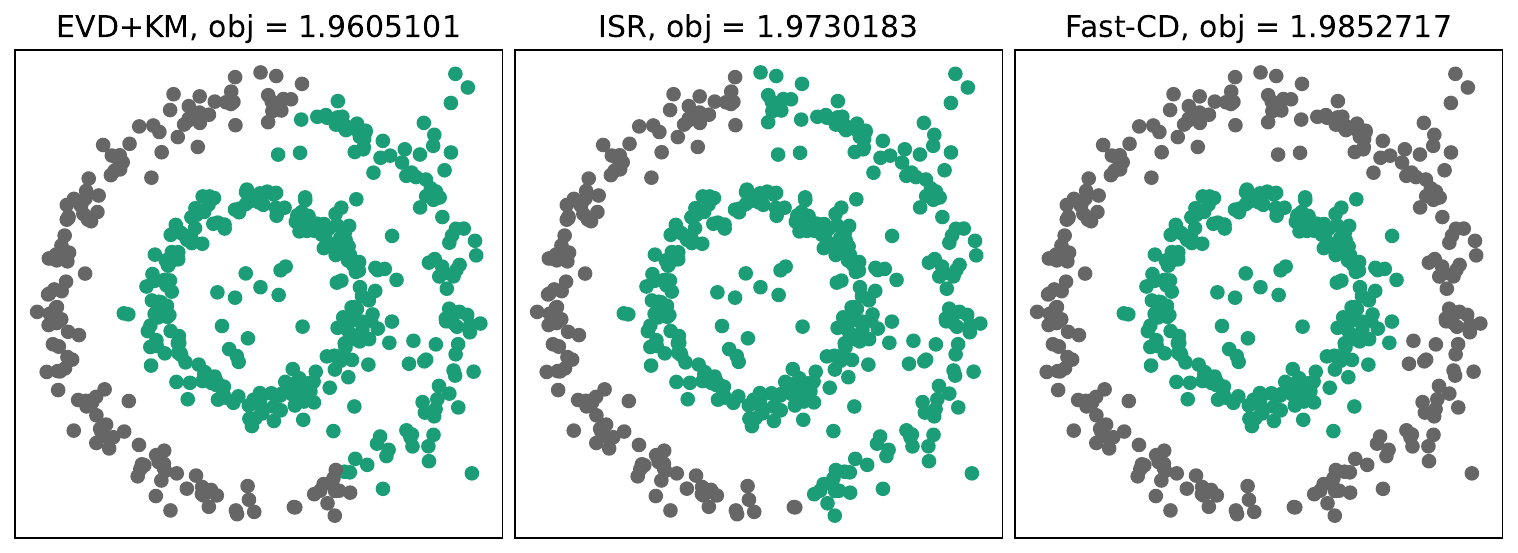}\label{fig:toy_100}}
	\\
	\subfloat[150 noise points. All competitors fail to cluster the outer circle,
		but Fast-CD still achieves the best N-Cut objective.]
	{\includegraphics[width=\linewidth]{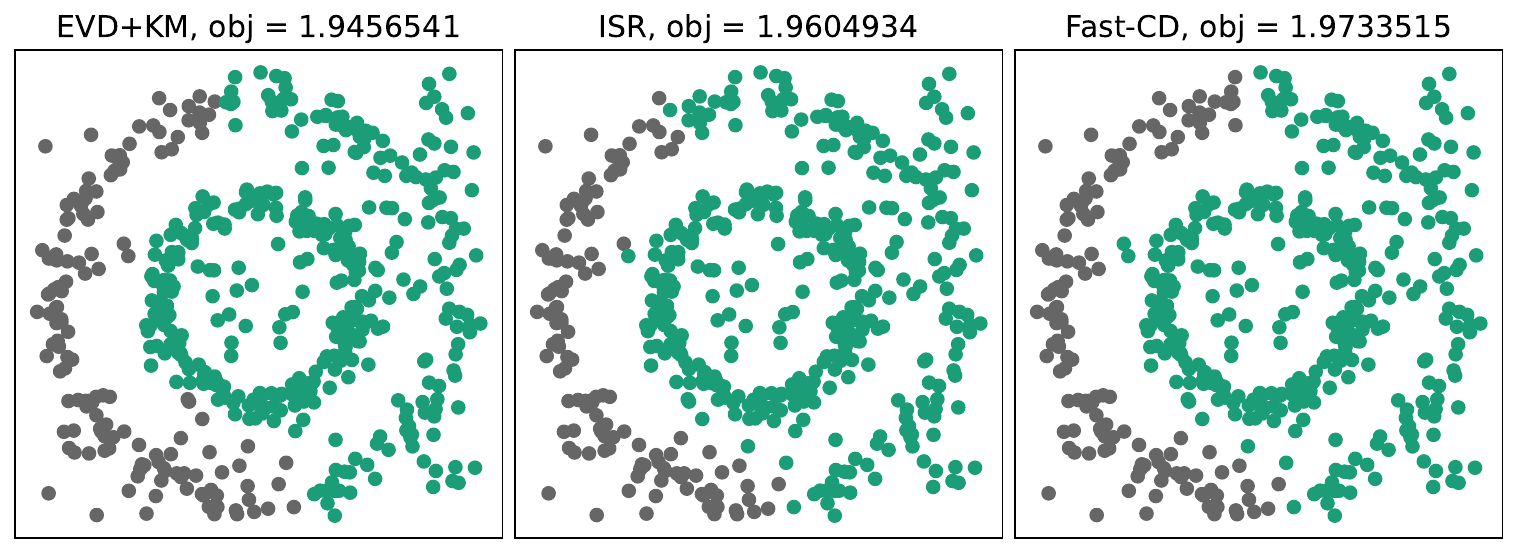}}
	\caption{Clustering results for noisy 2-circle data. Each circle contains 200
		points. Different numbers of extra uniform noise points are added.}
	\label{fig:toy}
\end{figure}

To demonstrate the importance of finding a more accurate solution, we follow
\cite{DBLP:journals/pr/ChangY08} to evaluate the performance on the
2-circle dataset, where uniform noise is gradually added. The results are
plotted in \cref{fig:toy}. We observe that Fast-CD consistently achieved the
largest N-Cut objective~\eqref{N-cut-max} with different noisy levels.
Especially, EVD+KM and ISR misclustered half of the outer circle in
\cref{fig:toy_100}, while the solution with a larger objective value found by
Fast-CD correctly clustered both circles.

\subsubsection{Image Segmentation}
\begin{figure}[t]
	\centering
	\subfloat[Input]{\includegraphics[width=0.22\linewidth]{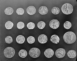}}
	\hfill
	\subfloat[EVD+KM, \texttt{obj}=$24.9698$,
		time=\qty{1.2351}{\second}]{\includegraphics[width=0.22\linewidth]{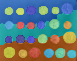}}
	\hfill
	\subfloat[ISR, \texttt{obj}=$24.9888$,
		time=\qty{18.2363}{\second}]{\includegraphics[width=0.22\linewidth]{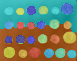}}
	\hfill
	\subfloat[Fast-CD, \texttt{obj}=$\mathbf{24.9969}$,
		time=\textbf{\qty[text-series-to-math]{0.1452}{\second}}]{\includegraphics[width=0.22\linewidth]{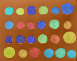}}
	\caption{Image segmentation results of different N-Cut solvers. The image
		consists of 24 coins outlined against a darker background thus 25 clusters
		in total. Fast-CD outperforms other competitors in three aspects: 1)
		obtained the largest N-Cut objective value; 2) produced notably the best
		segmentations; 3) finished the task in the shortest time.}%
	\label{fig:segmentation}
\end{figure}
Image segmentation is the task of labeling the pixels of objects of interest in
an image. A graph is first created from the voxel-to-voxel difference on an image,
and graph cuts are applied to break this image into multiple partly-homogeneous
regions. \cite{shi2000normalized,stella2003multiclass} are ancestor works
applying N-Cut to image segmentation.

Our experiment is derived from an example of
scikit-learn\footnote{\url{https://scikit-learn.org/stable/auto_examples/cluster/plot_coin_segmentation.html}}.
The image consists of 24 coins outlined against a darker background thus 25
clusters in total. We reuse their code to preprocess the image and construct
the graph. The input image and segmentation results are plotted in
\cref{fig:segmentation}. Obviously, both EVD+KM and ISR failed to properly
identify the background, while Fast-CD produced the most satisfactory, in which
all coins are properly separated from the background. In addition to the
segmentation quality, Fast-CD obtained the largest N-Cut objective value and
ran much faster than the competitors.

\subsection{Data Clustering}

\begin{table}[t]
	\centering
	\caption{Dataset statistics}
	\label{tab:datasets}
	\begin{tabular}{@{}cccc@{}}
		\toprule
		Name        & \# Samples & \# Features & \# Classes \\ \midrule
		German      & 1000       & 20          & 2          \\
		GTdb        & 750        & 21600       & 50         \\
		MNIST10     & 6996       & 784         & 10         \\
		MSRA50      & 1799       & 1024        & 12         \\
		Segment     & 2310       & 19          & 7          \\
		STL-10      & 13000      & 2048        & 10         \\
		UMist       & 575        & 644         & 20         \\
		Waveform-21 & 2746       & 21          & 3          \\ \bottomrule
	\end{tabular}
\end{table}

\begin{table*}[t]
	\centering
	\caption{Objective values and clustering performance of different N-Cut
		solvers on 8 real-world benchmark datasets, the best results are in bold.}
	\label{tab:ncut_obj}
	\setlength{\tabcolsep}{4.5pt}
	\begin{tabular}{@{}ccccccccccccc@{}}
		\toprule
		Algorithm & EVD+KM                      & ISR                        & Fast-CD                     & EVD+KM                          & ISR     & Fast-CD          & EVD+KM  & ISR     & Fast-CD          & EVD+KM          & ISR     & Fast-CD          \\ \midrule
		Dataset   & \multicolumn{3}{c}{German}  & \multicolumn{3}{c}{GTdb}   & \multicolumn{3}{c}{MNIST10} & \multicolumn{3}{c}{MSRA50}                                                                                                                         \\
		\cmidrule(r){1-1}\cmidrule(lr){2-4}\cmidrule(lr){5-7}\cmidrule(lr){8-10}\cmidrule(l){11-13}
		obj       & 1.9918                      & 1.9914                     & \textbf{1.9954}             & 15.3375                         & 15.6509 & \textbf{15.9180} & 8.5232  & 8.5785  & \textbf{8.6528}  & 11.6193         & 11.6190 & \textbf{11.6259} \\
		ACC       & 0.5420                      & 0.5470                     & \textbf{0.6070}             & 0.5240                          & 0.5347  & \textbf{0.5480}  & 0.6954  & 0.5768  & \textbf{0.7766}  & \textbf{0.5397} & 0.5331  & 0.5320           \\
		NMI       & 0.0026                      & 0.0031                     & \textbf{0.0077}             & 0.7021                          & 0.6873  & \textbf{0.7056}  & 0.6506  & 0.5962  & \textbf{0.7244}  & \textbf{0.6933} & 0.6805  & 0.6734           \\
		ARI       & 0.0054                      & 0.0068                     & \textbf{0.0294}             & \textbf{0.4060}                 & 0.3645  & 0.4002           & 0.5346  & 0.4490  & \textbf{0.6496}  & \textbf{0.4498} & 0.4362  & 0.4159           \\
		\midrule
		Dataset   & \multicolumn{3}{c}{Segment} & \multicolumn{3}{c}{STL-10} & \multicolumn{3}{c}{UMist}   & \multicolumn{3}{c}{Waveform-21}                                                                                                                    \\
		\cmidrule(r){1-1}\cmidrule(lr){2-4}\cmidrule(lr){5-7}\cmidrule(lr){8-10}\cmidrule(l){11-13}
		obj       & 6.8729                      & 6.9204                     & \textbf{6.9272}             & 9.2781                          & 9.2636  & \textbf{9.3206}  & 12.4520 & 12.4131 & \textbf{12.4825} & 2.7477          & 2.7489  & \textbf{2.7652}  \\
		ACC       & \textbf{0.5043}             & 0.4983                     & 0.4965                      & 0.9312                          & 0.8860  & \textbf{0.9433}  & 0.4191  & 0.4191  & \textbf{0.4730}  & 0.5215          & 0.5229  & \textbf{0.5663}  \\
		NMI       & \textbf{0.5088}             & 0.4971                     & 0.4875                      & 0.8783                          & 0.8456  & \textbf{0.8929}  & 0.6302  & 0.6205  & \textbf{0.6765}  & 0.3706          & 0.3709  & \textbf{0.3795}  \\
		ARI       & \textbf{0.3738}             & 0.3583                     & 0.3403                      & 0.8574                          & 0.7870  & \textbf{0.8813}  & 0.3240  & 0.3166  & \textbf{0.3668}  & 0.2530          & 0.2538  & \textbf{0.2778}  \\ \bottomrule
	\end{tabular}
\end{table*}

In this section, we employ 8 real-world benchmark datasets of varying numbers of
samples and features that cover a wide range to evaluate the clustering
performance. \texttt{German}, \texttt{Segment} and \texttt{Waveform-21} are
grabbed from the UCI Machine Learning Repository~\cite{asuncion2007uci}.
\texttt{MNIST10} is a subset of the famous handwritten digit
dataset~\cite{lecun1998gradient}. \texttt{STL-10} is an object
dataset~\cite{coates2011analysis} and we use the features extracted by a
ResNet50~\cite{he2016deep} convolution neural network. Their statistics
characters are summarized in \cref{tab:datasets}. We employ the self-tuning
spectral clustering strategy~\cite{DBLP:conf/nips/Zelnik-ManorP04} to construct
the initial graphs.

Apart from the N-Cut objective value, we also employ three widely adopted
metrics to quantify the clustering performance, including clustering accuracy
(ACC), normalized mutual information (NMI)~\cite{strehl2002cluster} and
adjusted Rand index (ARI)~\cite{hubert1985comparing}.

\begin{table}[t]
	\centering
	\caption{Averaged execution time (seconds) of 10 individual runs.}
	\label{tab:time}
	\addtolength{\tabcolsep}{-4pt}
	\renewcommand{\arraystretch}{1.1}
	\begin{adjustbox}{max width=\linewidth}
		\begin{threeparttable}
			\begin{tabular}{@{}cccccc@{}}
				\toprule
				\multirow{2}{*}{\diagbox[trim=lr]{Dataset}{Metric}} & \multicolumn{2}{c}{EVD+KM}     & \multicolumn{2}{c}{ISR} & Fast-CD                                                                           \\
				\cmidrule(lr){2-3}\cmidrule(lr){4-5}\cmidrule(l){6-6}
				                                                    & mean±std                       & Accel                   & mean±std                & Accel         & mean±std                                \\
				\midrule
				German                                              & $0.037\colorpm{\mathbf{0.00}}$ & $7.1\times$             & $0.201\colorpm{0.03}$   & $39.0\times$  & $\mathbf{0.005}\colorpm{\mathbf{0.00}}$ \\
				GTdb                                                & $0.082\colorpm{0.04}$          & $7.0\times$             & $0.353\colorpm{0.03}$   & $30.1\times$  & $\mathbf{0.012}\colorpm{\mathbf{0.00}}$ \\
				MNIST10                                             & $2.702\colorpm{0.21}$          & $10.1\times$            & $43.998\colorpm{0.68}$  & $164.2\times$ & $\mathbf{0.268}\colorpm{\mathbf{0.01}}$ \\
				MSRA50                                              & $0.138\colorpm{0.02}$          & $8.9\times$             & $0.898\colorpm{0.07}$   & $57.8\times$  & $\mathbf{0.016}\colorpm{\mathbf{0.00}}$ \\
				Segment                                             & $0.212\colorpm{0.02}$          & $11.0\times$            & $1.726\colorpm{0.08}$   & $89.8\times$  & $\mathbf{0.019}\colorpm{\mathbf{0.00}}$ \\
				STL-10                                              & $12.515\colorpm{0.33}$         & $21.8\times$            & $224.203\colorpm{9.47}$ & $390.7\times$ & $\mathbf{0.574}\colorpm{\mathbf{0.03}}$ \\
				UMist                                               & $0.032\colorpm{0.02}$          & $5.0\times$             & $0.099\colorpm{0.01}$   & $15.2\times$  & $\mathbf{0.007}\colorpm{\mathbf{0.00}}$ \\
				Waveform-21                                         & $0.249\colorpm{0.02}$          & $5.7\times$             & $2.722\colorpm{0.06}$   & $62.9\times$  & $\mathbf{0.043}\colorpm{\mathbf{0.00}}$ \\ \bottomrule
			\end{tabular}
			\begin{tablenotes}
				\item \texttt{Accel} denotes the acceleration ratio of Fast-CD, the fastest
				and most stable results are in bold.
			\end{tablenotes}
		\end{threeparttable}
	\end{adjustbox}
\end{table}

\subsubsection{Clustering with the Same Initialization}

We run all comparison methods with the same initial cluster assignments and
report the results in \cref{tab:ncut_obj}. The following observations are
obtained from the results:
\begin{itemize}
	\item Our proposed Fast-CD solver consistently obtained the largest objective
	      value of N-Cut (\cref{N-cut-max}), which means that our solver is able to
	      find a better local minimum compared with the ordinary EVD+KM solver and
	      ISR.
	\item Fast-CD obtained the best clustering performance \wrt all three metrics
	      on 5 out of 8 datasets. This indicates that achieving better solutions to
	      the N-Cut problem is indeed beneficial for boosting the clustering
	      performance.
	\item Fast-CD demonstrates its superb performance on \texttt{MNIST10}
	      dataset, with \textbf{+0.07} of ACC, NMI, and ARI compared to
	      the second-best method, which is a huge improvement.
	\item Despite larger objective values, both Fast-CD and ISR failed to surpass
	      EVD+KM on \texttt{Segment} dataset. In addition,
	      Fast-CD obtained a larger objective value than ISR but produced
	      worse cluster assignments regarding ACC, NMI, and ARI. We argue that they belong
	      to the rare cases where the N-Cut model failed to capture the
	 clustering relationships on certain datasets precisely, and thus better solutions of
	      N-Cut lead to worse clustering performances. The issue is attributed to the
	      design of the clustering model thus orthogonal to this work.
\end{itemize}

\begin{table}[t]
	\centering
	\caption{N-Cut objective values and NMI of Fast-CD initialized via EVD+KM.}
	\label{tab:ncut_init}
	\addtolength{\tabcolsep}{-3pt}
	\begin{tabular}{@{}ccccc@{}}
		\toprule
		\multirow{2}{*}{\diagbox[trim=lr]{Dataset}{Metric}} & \multicolumn{2}{c}{obj} & \multicolumn{2}{c}{NMI}                                     \\
		\cmidrule(lr){2-3} \cmidrule(l){4-5}
		                                                    & start                   & end                     & start           & end             \\
		\midrule
		German                                              & 1.991767545             & \textbf{1.992184158}    & 0.0026          & \textbf{0.0028} \\
		GTdb                                                & 15.33753865             & \textbf{15.95620542}    & \textbf{0.7021} & 0.7004          \\
		MNIST10                                             & 8.523162624             & \textbf{8.616931822}    & 0.6506          & \textbf{0.6784} \\
		MSRA50                                              & 11.61930657             & \textbf{11.63278001}    & 0.6933          & \textbf{0.6955} \\
		Segment                                             & 6.872888591             & \textbf{6.888504946}    & \textbf{0.5088} & 0.5073          \\
		STL-10                                              & 9.278119471             & \textbf{9.325558444}    & 0.8783          & \textbf{0.8886} \\
		UMist                                               & 12.45199326             & \textbf{12.59905529}    & 0.6302          & \textbf{0.6396} \\
		Waveform-21                                         & 2.747671034             & \textbf{2.757772927}    & 0.3706          & \textbf{0.3741} \\ \bottomrule
	\end{tabular}
\end{table}

\begin{figure}[t]
	\centering
	\subfloat[MNIST10]{\includegraphics[width=0.5\linewidth]{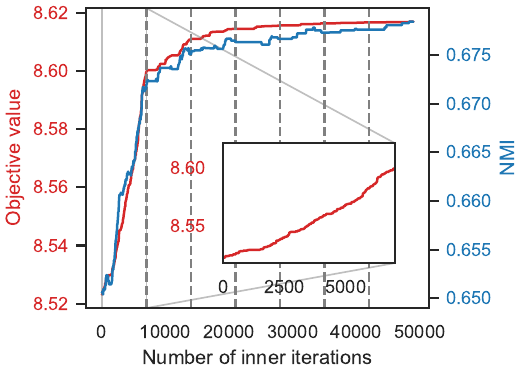}}
	\subfloat[STL-10]{\includegraphics[width=0.5\linewidth]{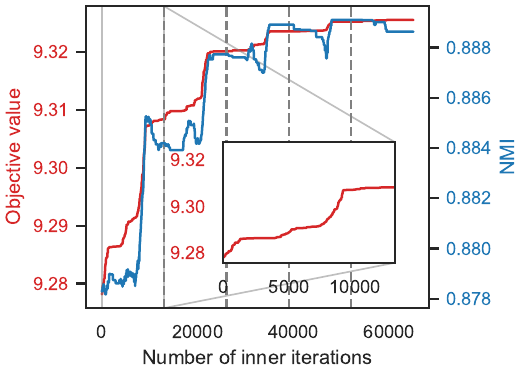}}
	\caption{N-Cut objective and NMI curves of Fast-CD initialized by EVD+KM
		until convergent.
		Gray dashes separate different outer iterations. The zoomed insets plot the
		objective values within the first outer iteration, where the rows of
		$\mathbf{Y}$ are updated sequentially.}%
	\label{fig:ncut_init}
\end{figure}

\subsubsection{Execution Time}

To demonstrate the efficiency of the proposed Fast-CD solver, we compare the
execution time of all three solvers on the aforementioned 8 datasets. We focus
on the actual clustering operations so that boilerplate parts including loading
datasets and initialization are not recorded. All executions are repeated 10
times and we report the mean and standard deviation of the execution time in
\cref{tab:time}. Additionally, we present the acceleration ratio of Fast-CD as
$\texttt{Accel} = \mathrm{mean}_{\mathrm{comparison~method}} /
	\mathrm{mean}_{\mathrm{Fast-CD}}$ for clear exposition.

Obviously, Fast-CD runs extremely fast: finished clustering
in 1s on all datasets, and consistently outperforms others by a large margin.
The superiority is especially significant on large-scale datasets such as
\texttt{MNIST10} and \texttt{STL-10}. This is because the time complexity of
Fast-CD is just linear to $n$ while others are cubic or higher. The
acceleration becomes more noticeable as $n$ becomes larger, which largely
enhances the practicality in real-world applications. On the other hand, the
standard deviations of Fast-CD are much smaller than the competitors and close
to zero, which means the execution is very stable.

Despite ISR being theoretically proven to perform better than
EVD+KM~\cite{stella2003multiclass,huang2013spectral,chen2017scalable} and
empirically verified in \cref{tab:ncut_obj}, it's much more computationally
expensive than EVD+KM according to \cref{tab:time} and can lose the
effectiveness-efficiency trade-off. In reverse, Fast-CD is faster and better
comparing with both EVD+KM and ISR.

\subsubsection{Initialize Fast-CD with EVD+KM}%

In this section, we use the cluster assignments learned by EVD+KM to initialize
the proposed Fast-CD solver. The objective value of N-Cut and NMI are presented
in \cref{tab:ncut_init}. We observe that Fast-CD is always able to increase
the N-Cut objective on top of the ordinary EVD+KM predictions, \ie, Fast-CD is able
to refine the clustering assignments found by EVD+KM and converge to a local
minimum of the original N-Cut objective.
Along with the increasing objective value, Fast-CD further improved the
clustering performance of EVD+KM predictions in most cases.

We plot N-Cut objective and NMI curves of Fast-CD in \cref{fig:ncut_init} to
further demonstrate how they evolve until Fast-CD converges, in which the
results are obtained from two large datasets \texttt{MNIST10} and
\texttt{STL-10}. The objective value increases monotonically and NMI tends to
improve with the objective. Fast-CD converged after 6 iterations on both
datasets, indicating its superb efficiency.

\subsection{Compare with Graph-based Clustering Methods}

In this section, we additionally compare the performance of the N-Cut model with
other popular graph-based clustering models, including CLR-$\ell_1$,
CLR-$\ell_2$~\cite{nie2016constrained}, SBMC~\cite{chen2017self} and
EBMC~\cite{chen2020enhanced}. As we have already verified that Fast-CD is the
most efficient and effective N-Cut solver, EVD+KM and ISR results are omitted.
We present the NMIs and acceleration ratios in \cref{tab:nmi_graph_based}. It
can be shown that recent works such as SBMC and EBMC have comparable while
slightly better clustering performance than Fast-CD, indicating the bottleneck
of the N-Cut model. However, considering the absolute advantage of Fast-CD in terms
of execution time, it's still applicable and preferred for large-scale data and
time-sensitive scenarios.

\begin{table}[t]
	\centering
	\caption{NMI and time comparison among graph-based clustering methods.}
	\label{tab:nmi_graph_based}
	\addtolength{\tabcolsep}{-4pt}
	\renewcommand{\arraystretch}{1.3}
	\begin{adjustbox}{max width=\linewidth}
		\begin{threeparttable}
			\begin{tabular}{@{}cccccccccc@{}}
				\toprule
				\multirow{2}{*}{\diagbox[trim=lr]{Dataset}{Method}} & \multicolumn{2}{c}{CLR-$\ell_1$} & \multicolumn{2}{c}{CLR-$\ell_2$} & \multicolumn{2}{c}{SBMC} & \multicolumn{2}{c}{EBMC} & Fast-CD                                                                             \\
				\cmidrule(lr){2-3} \cmidrule(lr){4-5} \cmidrule(lr){6-7} \cmidrule(lr){8-9} \cmidrule(l){10-10}
				                                                    & NMI                              & Accel                            & NMI                      & Accel                    & NMI             & Accel         & NMI             & Accel         & NMI             \\
				\midrule
				German                                              & 0.0029                           & $230.9\times$                    & 0.0029                   & $247.4\times$            & 0.0074          & $4.2\times$   & \textbf{0.0077} & $21.9\times$  & \textbf{0.0077} \\
				GTdb                                                & 0.6536                           & $75.5\times$                     & 0.6539                   & $57.9\times$             & \textbf{0.7386} & $22.8\times$  & \textbf{0.7386} & $41.3\times$  & 0.7056          \\
				MNIST10                                             & 0.6539                           & $>500\times$                     & 0.6549                   & $>500\times$             & 0.6456          & $29.2\times$  & \textbf{0.7318} & $95.6\times$  & 0.7244          \\
				MSRA50                                              & 0.6101                           & $436.7\times$                    & 0.6275                   & $442.9\times$            & \textbf{0.6763} & $12.8\times$  & 0.6715          & $102.9\times$ & 0.6734          \\
				Segment                                             & 0.2150                           & $159.9\times$                    & 0.2150                   & $133.1\times$            & 0.4684          & $113.5\times$ & 0.4819          & $80.0\times$  & 0.4875          \\
				STL-10                                              & 0.8872                           & $>500\times$                     & 0.6778                   & $>500\times$             & 0.8343          & $39.3\times$  & 0.8919          & $138.4\times$ & \textbf{0.8929} \\
				UMist                                               & \textbf{0.8411}                  & $136.5\times$                    & 0.8332                   & $69.1\times$             & 0.6980          & $15.1\times$  & 0.6825          & $23.4\times$  & 0.6765          \\
				Waveform-21                                         & 0.3701                           & $>500\times$                     & 0.3785                   & $>500\times$             & \textbf{0.4172} & $57.8\times$  & 0.3811          & $47.8\times$  & 0.3795          \\ \bottomrule
			\end{tabular}
			\begin{tablenotes}
				\item \texttt{Accel} denotes the acceleration ratio of Fast-CD.
			\end{tablenotes}
		\end{threeparttable}
	\end{adjustbox}
\end{table}

\section{Conclusion}

We propose a novel, effective and efficient solver based on the famous
coordinate descent method for Normalized-Cut, namely Fast-CD. Unlike
spectral-based approaches of $\bigO(n^3)$ time complexity, the whole time
complexity of our solver is just $\bigO(|E|)$. To avoid reliance on random
initialization which brings uncertainties to clustering, an efficient
initialization method that gives deterministic outputs is also designed.
Extensive experiments strongly support the superiority of our proposal.


\bibliographystyle{IEEEtran}
\bibliography{IEEEabrv,ref}

\ifCLASSOPTIONcaptionsoff
	\newpage
\fi



%




\end{document}